%% file: neurips_2026.tex
\documentclass[table]{article}

\PassOptionsToPackage{authoryear, round}{natbib}

\usepackage[preprint]{neurips_2026}

\usepackage[utf8]{inputenc} 
\usepackage[T1]{fontenc}    
\usepackage{url}            
\usepackage{booktabs}       
\usepackage{amsfonts}       
\usepackage{nicefrac}       
\usepackage{microtype}      
\usepackage[colorlinks,
            linkcolor=red,
            anchorcolor=blue,
            citecolor=green
            ]{hyperref}

\usepackage{xcolor}
\usepackage{natbib}
\setcitestyle{authoryear,round}
\usepackage{graphicx}
\usepackage{amssymb}
\usepackage{multirow}
\usepackage{arydshln}
\usepackage[listings]{tcolorbox}
\usepackage{footmisc}
\usepackage{enumitem}
\usepackage{array}
\usepackage{amsmath}
\usepackage{wrapfig}

\usepackage{algorithm}
\usepackage{algpseudocode}
\usepackage{eso-pic}

\newcommand{\longcatlogoposfirst}{\AtPageUpperLeft{\hspace{37.7mm}\raisebox{-26.3mm}{\includegraphics[height=9mm]{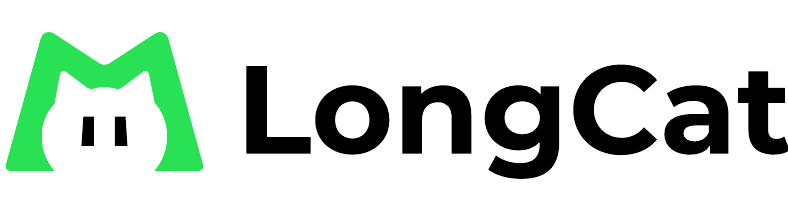}}}}

\AddToShipoutPictureFG{%
  \ifnum\value{page}=1\relax
    \longcatlogoposfirst
  \fi
}

\usepackage{inconsolata}
\usepackage{tabularx}
\usepackage{makecell}
\usepackage{mathrsfs}
\usepackage{amsthm}
\usepackage{subfigure}
\usepackage{caption}
\usepackage{soul}
\usepackage{pifont}

\definecolor{lightgreen}{rgb}{0.85,1.0,0.85}
\sethlcolor{lightgreen}

\definecolor{+}{RGB}{35, 225, 35}
\definecolor{-}{RGB}{224, 25, 25}

\definecolor{mycyan}{RGB}{0, 158, 150}
\definecolor{algcomment}{RGB}{70, 130, 180}
\algrenewcommand{\algorithmiccomment}[1]{\hfill\textcolor{algcomment}{\(\triangleright\) #1}}
\newtcolorbox{conclusionbox}{
  colback=mycyan!10,
  colframe=white,
  boxrule=0pt,
  arc=4pt,
  left=3pt,
  right=3pt,
  top=3pt,
  bottom=3pt,
}

\newcommand{\ie}{\emph{i.e., }}
\newcommand{\eg}{\emph{e.g., }}

\newcommand{\cf}{\emph{cf. }}

\title{Skill1: Unified Evolution of Skill-Augmented Agents via Reinforcement Learning}

\author{
  \textbf{Yaorui Shi$^{1,2,*}$},
  \textbf{Yuxin Chen$^{2,3,*}$},
  \textbf{Zhengxi Lu$^{2,4}$},
  \textbf{Yuchun Miao$^{2,5}$},
  \textbf{Shugui Liu$^{1}$},
  \\
  \textbf{Qi Gu$^{2,\dagger}$},
  \textbf{Xunliang Cai$^{2}$},
  \textbf{Xiang Wang$^{1}$},
  \textbf{An Zhang$^{1,\dagger}$} \\
  \vspace{-2mm} \\
  $^{1}$University of Science and Technology of China,
  $^{2}$Meituan, \\
  $^{3}$National University of Singapore,
  $^{4}$Zhejiang University, \\
  $^{5}$Wuhan University,
  $^{*}$Equal contribution.\\
  $^{\dagger}$Corresponding authors: \texttt{guqi03@meituan.com}, \texttt{an\_zhang@ustc.edu.cn}
}

\begin{document}

\maketitle

\input{chapters/0_abstract}
\input{chapters/1_intro}
\input{chapters/2_preliminary}
\input{chapters/3_method}
\input{chapters/4_experiments}
\input{chapters/5_conclusion}

\newpage
\bibliographystyle{unsrtnat}
\bibliography{custom}

\input{chapters/99_appendix}

\end{document}

%% file: chapters/0_abstract.tex
\begin{abstract}
A persistent skill library allows language model agents to reuse successful strategies across tasks.
Maintaining such a library requires three coupled capabilities. The agent selects a relevant skill, utilizes it during execution, and distills new skills from experience.
Existing methods optimize these capabilities in isolation or with separate reward sources, resulting in partial and conflicting evolution.
We propose Skill1, a framework that trains a single policy to co-evolve skill selection, utilization, and distillation toward a shared task-outcome objective.
The policy generates a query to search the skill library, re-ranks candidates to select one, solves the task conditioned on it, and distills a new skill from the trajectory.
All learning derives from a single task-outcome signal. Its low-frequency trend credits selection and its high-frequency variation credits distillation.
Experiments on ALFWorld and WebShop show that Skill1 outperforms prior skill-based and reinforcement learning baselines.
Training dynamics confirm the co-evolution of the three capabilities, and ablations show that removing any credit signal degrades the evolution.
Our code is available at \url{https://github.com/AlphaLab-USTC/Skill1}.
\end{abstract}

%% file: chapters/1_intro.tex
\section{Introduction}
\label{sec:introduction}

Reinforcement learning (RL)~\citep{sutton-barto,ppo,deepseekmath} has become an important paradigm for training large language model (LLM) agents that interact with complex environments~\citep{deepseekr1,qwen2.5,longcat,llama,alfworld,webshop,agentgym}.
Standard RL training treats each task as an isolated episode, where the strategies that lead to success are absorbed only implicitly into the policy parameters and cannot be explicitly reused on future tasks.
A natural solution is to augment agents with a persistent skill library that accumulates reusable strategies from past experience, so that the agent can draw on previously successful approaches instead of solving every task from scratch~\citep{voyager,expel,skillrl,retroagent,comp-rl,skill0}.
The workflow of these skill-augmented agents can be organized around a three-stage lifecycle~\citep{sok-agentic-skills}: skill selection, where the agent selects a relevant skill from the library for the current task; skill utilization, where the agent executes guided by the selected skill; and skill distillation, where the agent derives new reusable skills from the trajectories.

Existing methods have advanced each stage through RL, improving how agents select skills~\citep{memskill,skillorchestra,arise,evolver}, utilize them~\citep{skillrl,comp-rl,retroagent,arise,sage}, and distill reusable knowledge~\citep{retroagent,sage,comp-rl,evolver}.
Yet two fundamental questions remain open.
\textbf{(1) How can an agent evolve all three capabilities simultaneously?}
Existing methods apply policy updates to only a subset of the lifecycle, leaving at least one capability unoptimized, leading to optimization bottlenecks~\citep{skillrl,comp-rl,retroagent,sage}.
For example, a policy that has learned to use skills well still underperforms if it keeps routing to sub-optimal ones.
\textbf{(2) How can the three capabilities co-evolve toward a shared objective?}
Prior designs draw the rewards from different sources~\citep{arise,retroagent,comp-rl}.
For example, one capability may receive task-outcome reward while another relies on an auxiliary signal such as self-assessed quality or heuristic matching scores.
Since the three capabilities jointly determine task success, optimizing them with inconsistent signals creates conflicting pressures.

\begin{figure}[t]
    \centering
    \includegraphics[width=0.9\linewidth]{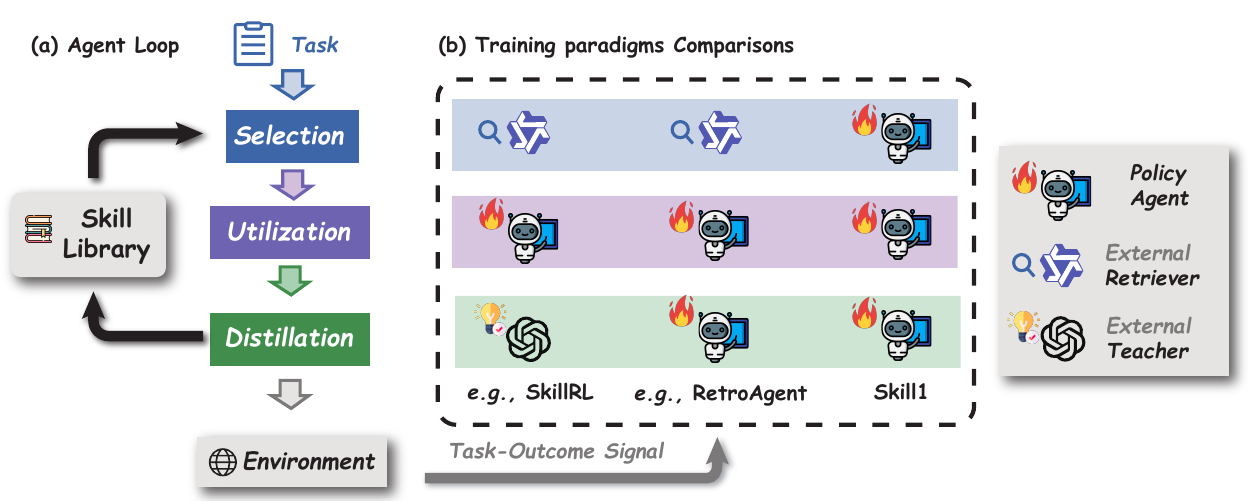}
    \caption{
    \textbf{Training paradigms for skill-augmented agents.}
    (a) The skill-augmented agent loop consists of selection, utilization, and distillation.
    (b) Prior methods delegate some stages to external modules without policy gradients (\eg freezes selection or uses an external teacher for distillation). Skill1 trains a single policy across all three stages with a shared task-outcome signal.
    }
    \vspace{-1em}
    \label{fig:teaser}
\end{figure}

We present Skill1, a framework that achieves unified evolution of skill-augmented agents by training a single policy to co-evolve skill selection, utilization, and distillation.
As illustrated in Figure~\ref{fig:teaser}, given a new task, the policy first generates a natural-language query to retrieve candidate skills from the library, and then re-ranks the retrieved candidates to select the best match.
The policy then performs multi-turn interaction with the environment conditioned on the top-ranked skill.
After execution, the policy distills reusable skills from the experience based on its rollouts.

We achieve co-evolution of all three capabilities through credit assignment on a single task-outcome signal $r(\tau)$.
The outcome directly measures how well the policy solves the current task and serves as the utilization reward.
To credit selection and distillation, we decompose this signal into its low-frequency trend and high-frequency variation.
The low-frequency trend is defined as the moving average of outcomes associated with each skill. This term reflects skill utility and guides the policy toward consistently effective skills.
The high-frequency variation is approximated with the deviation of the current outcome from the trend. This term captures whether a newly distilled skill improves upon the library's current boundary, and rewards the policy for producing useful skills.

We empirically evaluate Skill1 on ALFWorld~\citep{alfworld} and WebShop~\citep{webshop}.
Skill1 achieves 97.5\% success rate on ALFWorld, surpassing all other baseline skill-augmented agents.
Training dynamics confirm that selection precision, utilization success rate, and library quality improve simultaneously under the shared signal.
Ablations show that removing any single stage's credit-assignment signal degrades all three capabilities, providing evidence for their mutual dependence.

%% file: chapters/2_preliminary.tex
\section{Preliminary: LLM Agent with Skill Library}
\label{sec:preliminary}

\paragraph{Task formulation.}
We formulate the skill-augmented agent learning problem as a POMDP~\citep{pomdp-survey} $\mathcal{M} = (\mathcal{S}, \mathcal{A}, \mathcal{O}, T, \Omega, R, \gamma)$.
A state $S = (x, e, \mathcal{B})$ comprises a task instruction $x$ from dataset $\mathcal{D}$, the environment state $e$, and a persistent skill library $\mathcal{B} = \{s_1, s_2, \ldots\}$.
At each turn the agent selects an action $a \in \mathcal{A}$ to send to the environment.
The observation function $\Omega$ exposes a partial view $o_t = (x,\, e_t,\, z)$, where $z$ is the skill selected from $\mathcal{B}$ via a frozen encoder $\mathcal{E}$.
The overall training objective for the workflow can be defined as:
\begin{equation}
\label{eq:objective}
\max_\theta \; \mathbb{E}_{x \sim \mathcal{D},\, \tau \sim \pi_\theta(\cdot \mid x)}\bigl[r(\tau)\bigr],
\end{equation}
where $\pi_\theta$ is optimized with RL algorithms such as GRPO~\citep{deepseekmath} (\cf Appendix~\ref{app:grpo}).

\paragraph{Skills for LLM agents.}
\label{sec:skill_def}
A skill $s \in \mathcal{B}$ consists of a natural-language strategy $s.\text{strat}$ that describes how to act and a scenario description $s.\text{desc}$ that characterizes when the skill applies.
The agent maintains the skill library $\mathcal{B} = \{s_1, s_2, \ldots\}$ as it continuously explores the environment.
To reuse a skill, the agent generates its action conditioned on the skill strategy:
\begin{equation}
\label{eq:skill_conditioned}
a_t \sim \pi_\theta(\cdot \mid x,\, z.\text{strat},\, o_{\leqslant t}).
\end{equation}
To interact with a skill library, the agent selects skills from $\mathcal{B}$, utilizes them during execution (Eq.~\ref{eq:skill_conditioned}), and distills new skills back into $\mathcal{B}$.
In \S\ref{sec:method}, we show how to optimize all three stages jointly through a single policy, deriving every learning signal from the task outcome $r(\tau)$.

%% file: chapters/3_method.tex
\section{Method}
\label{sec:method}

\begin{figure}[t]
    \centering
    \includegraphics[width=\linewidth]{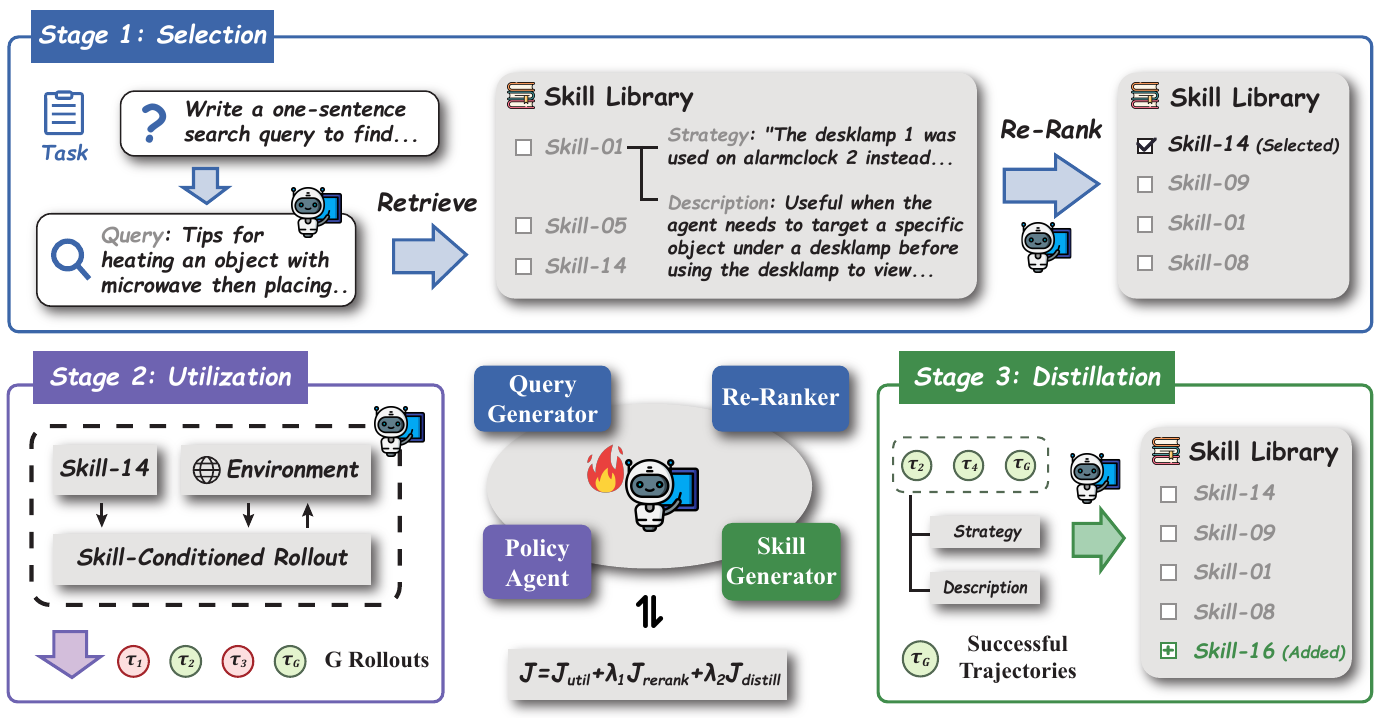}
    \caption{
    \textbf{Overview of the Skill1 framework.}
    (a) The policy generates a query and re-ranks retrieved candidates to select a skill.
    (b) The policy performs multi-turn interaction conditioned on the selected skill.
    (c) The policy reflects on the trajectory and distills a reusable skill.
    All learning signals are derived from the task-outcome $r(\tau)$ to achieve co-evolution of three capabilities.
    }
    \label{fig:framework}
\end{figure}

We introduce Skill1, a framework that trains a single policy $\pi_\theta$ to co-evolve skill selection, utilization, and distillation toward a shared task-outcome objective (Figure~\ref{fig:framework}).
We first describe the workflow (\S\ref{sec:workflow}), then derive all learning signals from the task outcome $r(\tau)$ (\S\ref{sec:reward}), and finally formulate the joint optimization objective (\S\ref{sec:optimization}).

\subsection{Agent Workflow}
\label{sec:workflow}

For each task $x \sim \mathcal{D}$, the policy $\pi_\theta$ performs three stages in sequence.
A complete trajectory takes the form $\tau = (q,\, z,\, a_1, o_1, \ldots, a_T, o_T,\, s_\text{new})$, where $q$ is the selection query, $z$ is the selected skill (or $\emptyset$), the action--observation pairs constitute the multi-turn interaction, and $s_\text{new}$ is the distilled skill.
The environment returns a terminal reward $r(\tau) \in \{0, 1\}$.
Prompt templates are in Appendix~\ref{app:prompts}.

\paragraph{Skill selection.}
Given a task $x$, the policy generates a natural-language query $q \sim \pi_\theta(\cdot \mid x)$ to search the skill library $\mathcal{B}$.
A frozen encoder $\mathcal{E}$ retrieves the top-$K$ candidates by semantic similarity:
\begin{equation}
\label{eq:selection}
\mathcal{B}_K = \operatorname{top\text{-}K}_{s \in \mathcal{B}}\; \operatorname{sim}\bigl(\mathcal{E}(q),\, \mathcal{E}(s.\text{desc})\bigr).
\end{equation}
The policy then re-ranks these candidates by generating a permutation $\sigma \sim \pi_\theta(\cdot \mid x, \mathcal{B}_K)$, and the top-ranked skill $z$ is provided for utilization.
Both query generation and re-ranking are produced by $\pi_\theta$, so selection is directly optimizable through the policy gradient.

\paragraph{Skill utilization.}
The policy interacts with the environment for up to $T$ turns conditioned on the selected skill:
$\tau \sim \pi_\theta(\cdot \mid x,\, z.\text{strat},\, o_{\leqslant t})$.
For each task, $G$ rollouts are sampled independently, each performing its own selection, utilization, and distillation.

\paragraph{Skill distillation.}
After each rollout, $\pi_\theta$ reflects on the trajectory to produce:
(i)~a reusable strategy $s_\text{new}.\text{strat} \sim \pi_\theta(\cdot \mid x, \tau)$ summarizing the approach, and
(ii)~a scenario description $s_\text{new}.\text{desc} \sim \pi_\theta(\cdot \mid x, \tau)$ characterizing when the skill applies.
A new skill is admitted to $\mathcal{B}$ only when $r(\tau) = 1$.
When the library reaches its capacity $|\mathcal{B}| = N_\text{max}$, the skill with the lowest retirement score $U(s) \cdot \log\bigl(n(s)\bigr)$ is removed, where $n(s)$ is the number of times $s$ has been selected.
This heuristic retires skills that are both low-utility and infrequently used while preserving well-tested high-utility skills.

\subsection{Reward Assignment}
\label{sec:reward}

Co-evolution requires that each capability receives targeted learning signals from the shared task outcome $r(\tau)$.
The challenge is that the three capabilities operate at different temporal scopes: utilization concerns the current episode, selection concerns which skills are consistently effective across episodes, and distillation concerns whether new experience improves upon what the library already covers.
We address this by decomposing $r(\tau)$ into its low-frequency trend and high-frequency variation, assigning credit to each capability without auxiliary models or additional rollouts.

\paragraph{Crediting utilization.}
The task outcome directly measures how well the policy executes with the given skill and serves as the utilization reward:
\begin{equation}
\label{eq:reward_utilize}
R^\text{util}_i \;=\; r(\tau_i).
\end{equation}

\paragraph{Crediting selection.}
Selection improves through two mechanisms.
First, the query $q$ is part of the rollout prefix and receives policy gradients through the utilization objective (Eq.~\ref{eq:utilize_obj}). Better queries retrieve better candidates and lead to higher $r(\tau)$, so query quality co-improves with task performance without a dedicated reward.

Second, re-ranking requires an explicit signal that reflects long-term skill quality rather than single-episode outcomes.
We maintain the trend of each skill as a per-skill utility score, updated after each rollout via exponential moving average:
\begin{equation}
\label{eq:utility}
U(s) \;\leftarrow\; (1 - \alpha) \cdot U(s) \;+\; \alpha \cdot r(\tau_i), \quad \forall\, s \in \mathcal{B}_K.
\end{equation}
We update all retrieved candidates rather than only the selected one, treating co-retrieval as evidence of relevance to the same task distribution.
The trend smooths out per-episode variance and accumulates each skill's long-term contribution.
We denote the best available utility as $\hat{U}_i = \max_{s \in \mathcal{B}_K^i} U(s)$, which serves as the library baseline for subsequent reward derivations.
The trend supervises re-ranking by rewarding the policy for producing a permutation $\sigma_i$ that agrees with the utility ordering. Here we use normalized discounted cumulative gain (NDCG) as the rubric:
\begin{equation}
\label{eq:reward_rerank}
R^\text{rerank}_i \;=\; \mathrm{NDCG}\bigl(\sigma_i,\; \operatorname{argsort}(-U(\mathcal{B}_K^i))\bigr).
\end{equation}

\paragraph{Crediting distillation.}
The ideal distillation signal would measure whether a newly distilled skill improves future task performance, but that future outcome is unavailable at training time.
We approximate it with the variation of the current outcome relative to the library's trend:
\begin{equation}
\label{eq:reward_distill}
R^\text{distill}_i \;=\; r(\tau_i) \;-\; \hat{U}_i,
\end{equation}
where $\hat{U}_i = \max_{s \in \mathcal{B}_K^i} U(s)$ is the highest trend among the retrieved candidates.
A positive variation indicates that the current experience surpasses what the library already covers, so the distilled skill is worth admitting. A negative variation discourages redundant distillation.

\subsection{Joint Optimization}
\label{sec:optimization}

\definecolor{commentblue}{RGB}{70, 130, 180}   
\definecolor{commentblue}{RGB}{100, 149, 237}  
\definecolor{commentblue}{RGB}{135, 170, 222}  
\definecolor{commentblue}{RGB}{65, 105, 225}   
\newcommand{\bigComment}[1]{\hfill\makebox[0.4\linewidth][l]{\textcolor{commentblue}{$\triangleright$ #1}}}
\algrenewcommand{\algorithmiccomment}[1]{%
  \hfill\makebox[0.34\linewidth][l]{\textcolor{commentblue}{$\triangleright$ #1}}%
}
\begin{algorithm*}[t]
\caption{Pseudo Code of Skill1}
\label{alg:skill1}
\small
\begin{algorithmic}[1]
\Require $\pi_\theta$, $\mathcal{B}$, $\mathcal{E}$, $K$, $G$, $\lambda_1$, $\lambda_2$, $\alpha$
\For{batch of $N$ tasks, each with $G$ rollouts}
    \For{sample $i = 1, \ldots, N \cdot G$} \bigComment{Agent workflow (Sec 3.1)}
        \State $q_i \gets \pi_\theta(x_i)$ \Comment{Skill selection: search}
        \State $\mathcal{B}_K \gets \operatorname{top\text{-}K}_{s \in \mathcal{B}}\, \operatorname{sim}\bigl(\mathcal{E}(q_i), \mathcal{E}(s.\text{desc})\bigr)$
        \State $\sigma_i \gets \pi_\theta(x_i, \mathcal{B}_K)$; \; $z_i \gets \mathcal{B}_K[\sigma_i(1)]$ \Comment{Skill Selection: re-rank}
        \State $\tau_i \sim \pi_\theta(\cdot \mid x_i,\, z_i.\text{strat})$ \Comment{Skill utilization}
        \State $(s_{\text{new},i}.\text{strat},\, s_{\text{new},i}.\text{desc}) \gets \pi_\theta(x_i, \tau_i)$ \Comment{Skill distillation}
    \EndFor
    \State $R^\text{util}_i \gets r(\tau_i)$; \; $\hat{U}_i \gets \max_{s \in \mathcal{B}_K^i} U(s)$ \bigComment{Reward assignment (Sec 3.2)}
    \State $R^\text{distill}_i \gets r(\tau_i) - \hat{U}_i$ \Comment{Variation as distillation credit}
    \State $R^\text{rerank}_i \gets \mathrm{NDCG}(\sigma_i,\, \operatorname{argsort}(-U(\mathcal{B}_K^i)))$ \Comment{Trend as selection credit}
    \State $U(s) \gets (1\!-\!\alpha)\, U(s) + \alpha\, r(\tau_i),\; \forall s \in \mathcal{B}_K^i$ \Comment{Update utility scores}
    \State Admit $s_{\text{new},i}$ to $\mathcal{B}$ if $r(\tau_i) = 1$ \Comment{Update skill library}
    \State $\theta \gets \theta + \nabla_\theta \bigl[\mathcal{J}^\text{util} + \lambda_1 \mathcal{J}^\text{rerank} + \lambda_2 \mathcal{J}^\text{distill}\bigr]$  \bigComment{Joint optimization (Sec 3.3)}
\EndFor
\end{algorithmic}
\end{algorithm*}

Each rollout $\tau_i$ is a concatenation of four generation segments produced by $\pi_\theta$: the selection query $q_i$, the re-ranking permutation $\sigma_i$, the action sequence $a_{1:T}$, and the distilled skill $s_{\text{new},i}$.
We assign each segment its own reward signal (\S\ref{sec:reward}) and optimize them jointly in a single gradient step using GRPO~\citep{deepseekmath} (\cf~Appendix~\ref{app:grpo}), which normalizes rewards within the $G$ rollouts of each task into group-relative advantages.

\paragraph{Utilization and query.}
The action tokens $a_{1:T}$ are conditioned on $(x_i, z_i)$ and optimized by the task outcome $R^\text{util}_i = r(\tau_i)$.
The query $q_i$ precedes the actions in the same sequence and receives gradients through the same objective:
\begin{equation}
\label{eq:utilize_obj}
\mathcal{J}^\text{util}(\theta) = \mathcal{J}_\text{GRPO}\bigl(\theta;\, \{\tau_1, \ldots, \tau_G\}, \{\hat{A}_1, \ldots, \hat{A}_G\}\bigr).
\end{equation}

\paragraph{Re-ranking.}
The permutation $\sigma_i$ is generated conditioned on the task $x_i$ and retrieved candidates $\mathcal{B}_K^i$, and reinforced by the ranking reward $R^\text{rerank}_i$.
Since different rollouts generate different queries, their retrieved candidate sets $\mathcal{B}_K^i$ differ, thus inner group comparison becomes meaningless.
We thus optimize each permutation independently with a REINFORCE-style~\citep{reinforce} objective:
\begin{equation}
\label{eq:rerank_obj}
\mathcal{J}^\text{rerank}(\theta) = \frac{1}{N \cdot G} \sum_i R^\text{rerank}_i \cdot \log \pi_\theta(\sigma_i \mid x_i, \mathcal{B}_K^i).
\end{equation}

\paragraph{Distillation.}
The distilled skill tokens $(s_{\text{new},i}.\text{strat},\, s_{\text{new},i}.\text{desc})$ are generated conditioned on the task $x_i$ and trajectory $\tau_i$, and reinforced by the variation $R^\text{distill}_i$.
Advantages $\hat{A}_i^\text{distill}$ are normalized separately from those of utilization since the two rewards measure different aspects of same outcomes:
\begin{equation}
\label{eq:distill_obj}
\mathcal{J}^\text{distill}(\theta) = \mathcal{J}_\text{GRPO}\bigl(\theta;\, \{s_{\text{new},1}, \ldots, s_{\text{new},G}\},\, \{\hat{A}_1^\text{distill}, \ldots, \hat{A}_G^\text{distill}\}\bigr).
\end{equation}

\paragraph{Total objective.}
All terms are combined in a single update:
\begin{equation}
\label{eq:total_objective}
\mathcal{J}(\theta) = \mathcal{J}^\text{util}(\theta) + \lambda_1\, \mathcal{J}^\text{rerank}(\theta) + \lambda_2\, \mathcal{J}^\text{distill}(\theta).
\end{equation}
The utility score $U(s)$ is updated non-parametrically via Eq.~\eqref{eq:utility}.
The full procedure is summarized in Algorithm~\ref{alg:skill1}. Training hyperparameter settings are in Appendix~\ref{app:hyperparams}.

%% file: chapters/4_experiments.tex
\section{Experiments}
\label{sec:experiments}

\subsection{Experimental Setup}
\label{sec:exp_setup}

\paragraph{Environments.}
We evaluate on ALFWorld~\citep{alfworld}, a text-based household environment requiring multi-step planning and object interaction, and WebShop~\citep{webshop}, an online-shopping simulator where agents search and purchase products matching user specifications.
We report success rate (\%) on the test split for both environments.

\paragraph{Training.}
For Skill1, the initial policy is Qwen2.5-7B-Instruct~\citep{qwen2.5} and the frozen encoder $\mathcal{E}$ is all-MiniLM-L6-v2~\citep{sentence-transformers}.
We train with GRPO under $G = 16$ and lr $= 1 \times 10^{-6}$. The skill library is initialized empty with capacity $|\mathcal{B}| \leqslant 5000$.
The training data uses the train split of the corresponding environments.
Full hyperparameters are in Appendix~\ref{app:hyperparams}.

\paragraph{Baselines.}
We compare three categories of methods in Table~\ref{tab:main_results}:
(1) training-free agents such as ReAct~\citep{react}, Reflexion~\citep{reflexion}, Mem0~\citep{mem0}, and ExpeL~\citep{expel};
(2) RL-trained methods without skills such as PPO~\citep{ppo}, RLOO~\citep{rloo}, GRPO~\citep{deepseekmath}, and GiGPO~\citep{gigpo};
and (3) RL-trained methods with skills such as EvolveR~\citep{evolver}, Mem0 and SimpleMem~\citep{simplemem} optimized with GRPO, SkillRL~\citep{skillrl}, and RetroAgent~\citep{retroagent}.
All baselines use the same base model Qwen2.5-7B-Instruct for fair comparison.

\subsection{Main Results}
\label{sec:main_results}

\begin{table}[t]
    \centering
    \small
    \caption{
    Main results on ALFWorld and WebShop (Success Rate, \%).
    \textbf{Bold} denotes best results; {\scriptsize\textcolor{mycyan}{$_{\uparrow}$}} indicates improvement over the previous best. ``Avg.'' stands for average success rate and ``Succ.'' stands for success rate.
    }
    \resizebox{\linewidth}{!}{
    \begin{tabular}{lccccccc|cc}
        \toprule
        & \multicolumn{7}{c|}{\textbf{ALFWorld (Success \%)}} & \multicolumn{2}{c}{\textbf{WebShop}} \\
        \cmidrule(lr){2-8} \cmidrule(lr){9-10}
        \textbf{Method} & Pick & Look & Clean & Heat & Cool & Pick2 & Avg. & Score & Succ. \\
        \midrule
        \rowcolor[gray]{0.92} \multicolumn{10}{c}{\textit{w/o Training}} \\
        Zero-Shot         & 33.4  & 21.6  & 19.3  & 6.9   & 2.8   & 3.2   & 14.8  & 26.4  & 7.8  \\
        ReAct~\citep{react}              & 48.5  & 35.4  & 34.3  & 13.2  & 18.2  & 17.6  & 31.2  & 46.2  & 19.5  \\
        Reflexion~\citep{reflexion}          & 62.0  & 41.6  & 44.9  & 30.9  & 36.3  & 23.8  & 42.7  & 58.1  & 28.8  \\
        Mem0~\citep{mem0}               & 54.0  & 55.0  & 26.9  & 36.4  & 20.8  & 7.7   & 33.6  & 23.9  & 2.0  \\
        ExpeL~\citep{expel}              & 21.0  & 67.0  & 55.0  & 52.0  & 71.0  & 6.0   & 46.3  & 30.9  & 11.2  \\
        \midrule
        \rowcolor[gray]{0.92} \multicolumn{10}{c}{\textit{RL-Trained w/o Skills}} \\
        PPO~\citep{ppo}                & 92.3  & 64.0  & 92.5  & 89.5  & 80.3  & 68.8  & 80.4  & 81.4  & 68.7  \\
        RLOO~\citep{rloo}              & 87.6  & 78.2  & 87.3  & 81.3  & 71.9  & 48.9  & 75.5  & 80.3  & 65.7  \\
        GRPO~\citep{deepseekmath}               & 90.8  & 66.1  & 89.3  & 74.7  & 72.5  & 64.7  & 77.6  & 79.3  & 66.1  \\
        GiGPO~\citep{gigpo}              & 97.7  & 82.7  & 98.8  & 83.7  & 89.3  & 79.2  & 90.8  & 84.4  & 72.8  \\
        \midrule
        \rowcolor[gray]{0.92} \multicolumn{10}{c}{\textit{RL-Trained w/ Skills}} \\
        EvolveR~\citep{evolver}            & 64.9  & 33.3  & 46.4  & 13.3  & 33.3  & 33.3  & 43.8  & 42.5  & 17.6  \\
        Mem0~\citep{mem0} w/ GRPO             & 78.1  & 54.8  & 56.1  & 31.0  & 65.0  & 26.9  & 54.7  & 58.1  & 37.5  \\
        SimpleMem~\citep{simplemem} w/ GRPO    & 89.5  & 36.3  & 60.0  & 50.0  & 64.9  & 26.3  & 62.5  & 67.8  & 46.9  \\
        SkillRL~\citep{skillrl}            & 97.9  & 71.4  & 90.0  & 90.0  & 95.5  & 87.5  & 89.9  & 85.2  & 72.7  \\
        RetroAgent~\citep{retroagent} & 97.9  & 90.9 & \textbf{99.2} & 92.9  & 85.3  & 91.0 & 94.9  & 88.9  & 82.3  \\
        \midrule
        \textbf{Skill1 (Ours)} & \textbf{100.0}{\scriptsize\textcolor{mycyan}{$_{\uparrow 2.1}$}} & \textbf{98.6}{\scriptsize\textcolor{mycyan}{$_{\uparrow 7.7}$}} & 97.3 & \textbf{99.2}{\scriptsize\textcolor{mycyan}{$_{\uparrow 6.3}$}} & \textbf{96.1}{\scriptsize\textcolor{mycyan}{$_{\uparrow 0.6}$}} & \textbf{96.0}{\scriptsize\textcolor{mycyan}{$_{\uparrow 5.0}$}} & \textbf{97.5}{\scriptsize\textcolor{mycyan}{$_{\uparrow 2.6}$}} & \textbf{89.7} & \textbf{82.9} \\
        \bottomrule
    \end{tabular}
    }
    \vspace{-0.5em}
    \label{tab:main_results}
\end{table}

Table~\ref{tab:main_results} presents the main results. We reproduce RetroAgent with the official implementation and borrow other baseline results from prior research~\citep{gigpo, skillrl, empg}. Skill1 results are averaged across three runs, and we report statistical analysis in Appendix~\ref{app:statistical_analysis}.

\textbf{Skill1 achieves the highest overall performance.}
On ALFWorld, Skill1 reaches 97.5\% average success rate, surpassing the previous best RetroAgent by 2.6 points and ranking first on five out of six task types.
The exception is Clean, where Skill1 falls slightly below strongest baseline RetroAgent, yet this gap is not statistically significant according to Appendix~\ref{app:statistical_analysis}.
On WebShop, Skill1 also demonstrates the best performance across all methods.

\textbf{An explicit skill library complements parameter-only RL.}
GiGPO, the strongest RL-only method, absorbs strategies implicitly into parameters and cannot explicitly reuse them across tasks.
Skill1 surpasses it by 6.7 points, with the largest gains on Look and Pick2 where composing multiple sub-procedures benefits most from reusable skills.

\textbf{Unified optimization outperforms methods that leave part of the lifecycle unoptimized.}
RetroAgent optimizes utilization and distillation with separate intrinsic rewards but provides no gradient signal for selection.
SkillRL freezes its selection mechanism after cold-start SFT.
Skill1 optimizes all three stages jointly through a single task-outcome signal. The comparison reveals a clear trend that agent performance increases with the degree of co-evolution.

\subsection{Analysis}
\label{sec:analysis}

\subsubsection{Ablation Study}
\label{sec:ablation}

We remove workflow components and zero out auxiliary objective weights to isolate each design choice.
All variants share the same base model and training budget. Results are reported in Table~\ref{tab:ablation}.

\begin{table}[t]
    \centering
    \small
    \caption{
    Ablation study on ALFWorld (Success Rate \%).
    Upper block ablates workflow components; lower block ablates training objectives.
    }
    \resizebox{0.7\linewidth}{!}{
    \begin{tabular}{l|cccccc|c}
        \toprule
         & Pick & Look & Clean & Heat & Cool & Pick2 & Avg. \\
        \midrule
        Skill1                   & 100.0 & 98.6 & 97.3 & 99.2 & 96.1 & 96.0 & 97.5 \\
        \midrule
        \quad w/o Select.             & 96.9 & 90.3 & 98.0 & 90.4 & 86.5 & 85.3 & 91.8 \\
        \quad w/o Distill.          & 97.4 & 88.5 & 98.1 & 96.1 & 87.6 & 89.5 & 92.4 \\
        \quad w/o Library               & 96.7 & 71.5 & 94.9 & 70.7 & 71.5 & 65.5 & 80.9 \\
        \midrule
        \quad w/ $\lambda_1{=}0$           & 99.5 & 80.5 & 98.8 & 100.0 & 90.6 & 84.9 & 94.0 \\
        \quad w/ $\lambda_2{=}0$           & 100.0 & 85.4 & 95.5 & 96.4 & 91.0 & 96.2 & 94.9 \\
        \quad w/ $\lambda_1{=}\lambda_2{=}0$ & 98.1 & 74.9 & 95.6 & 95.6 & 79.5 & 87.2 & 90.2 \\
        \bottomrule
    \end{tabular}
    }
    \label{tab:ablation}
\end{table}

\textbf{The skill library is the foundation, and distillation makes it effective.}
Removing the library entirely causes the largest drop, from 97.5\% to 80.9\%, with Heat and Pick2 losing over 28 points each.
These task types require composing multi-step sub-procedures that benefit most from reusable skills.
Removing distillation while keeping the library still reduces performance by 5.1 points.
Without distillation the library stores raw trajectories rather than condensed strategies, making selection noisier and reuse less effective.

\textbf{Selection loss propagates to downstream stages.}
Without selection the average drops by 5.7 points, concentrated on Heat and Pick2 where routing to the correct multi-step skill matters most.
Notably, this degradation occurs even though the utilization reward remains intact, showing that poor skill routing bottlenecks the entire pipeline regardless of the policy's solving ability.

\textbf{The two auxiliary objectives are complementary.}
Setting $\lambda_1{=}0$ or $\lambda_2{=}0$ individually reduces performance by 3.5 and 2.6 points respectively.
Removing both yields a sharper decline to 90.2\%, worse than removing each stage individually.
This gap shows that the signals benefit utilization beyond their direct targets, confirming that both signals are necessary to sustain full co-evolution.

\subsubsection{Co-evolution Dynamics}
\label{sec:training_dynamics}

\begin{figure}[t]
    \centering
    \includegraphics[width=\linewidth]{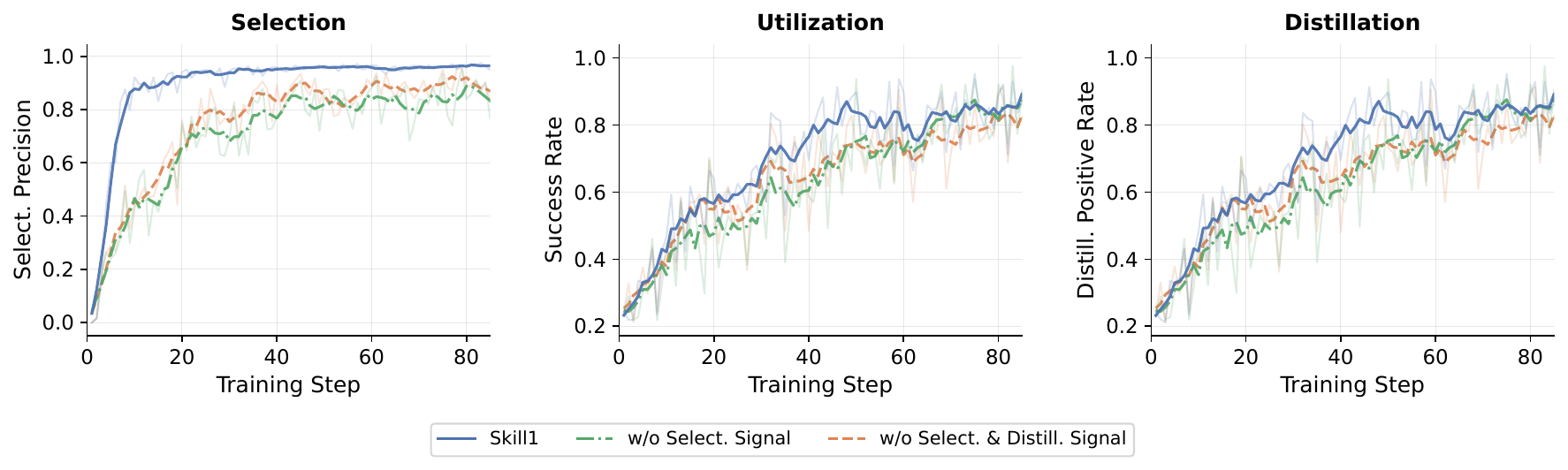}
    \caption{
    Training dynamics of the three capability metrics.
    Full Skill1 achieves fast and unified convergence across all stages.
    Removing selection signal (green) or both selection and distillation signals (orange) slows convergence of all capabilities.
    }
    \label{fig:coevolution}
\end{figure}

Figure~\ref{fig:coevolution} tracks three capability metrics across training:
(1) selection precision, the average skill utility scores $U(s)$;
(2) task-outcome reward $r(\tau)$ for utilization;
and (3) distillation positive rate, the fraction of new rollouts exceeding the average of retrieved ones $\hat{U}_i$.
We compare the full system against ablations that progressively remove credit-assignment signals.

\textbf{The three capabilities exhibit mutual reinforcement under unified training.}
Selection precision converges first, reaching 0.95 by step 20.
The resulting high-quality skill supply then accelerates the other two stages, with both utilization and distillation reaching 0.8 by step 60.
This sequential acceleration shows that improvements in one stage propagate forward through the lifecycle.

\textbf{Ablating any credit-assignment signal slows all three capabilities.}
Removing the selection signal reduces selection precision as expected, but also drags down utilization and distillation because the policy routes to sub-optimal skills more frequently.
Further removing distillation causes utilization scores to drop, even though it still receives its own direct reward.
This suggests that each signal contributes to the overall growing trend, which is a direct evidence of co-evolution.

\subsubsection{Selection and Distillation Quality}
\label{sec:skill_capability}

The previous section shows that capability metrics rise together. Here we examine the qualitative nature of that improvement: does the policy actually learn to select more relevant skills and distill higher-quality ones?

\begin{figure}[t]
    \centering
    \begin{minipage}[t]{0.52\textwidth}
        \centering
        \includegraphics[width=\linewidth]{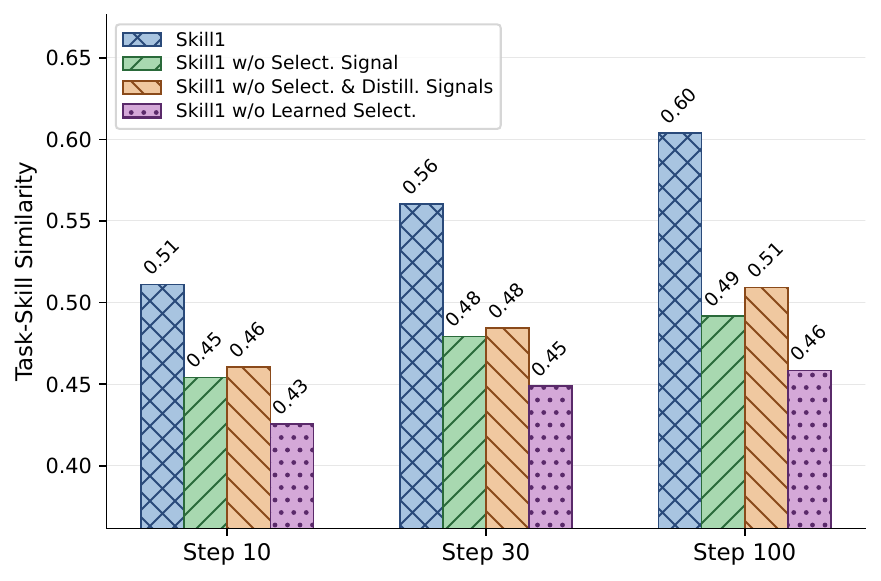}
        \caption{
        Task-skill similarity at three training checkpoints. The trend signal drives continuous improvement in selection quality.
        }
        \label{fig:retrieval_quality}
    \end{minipage}
    \hfill
    \begin{minipage}[t]{0.44\textwidth}
        \centering
        \includegraphics[width=\linewidth]{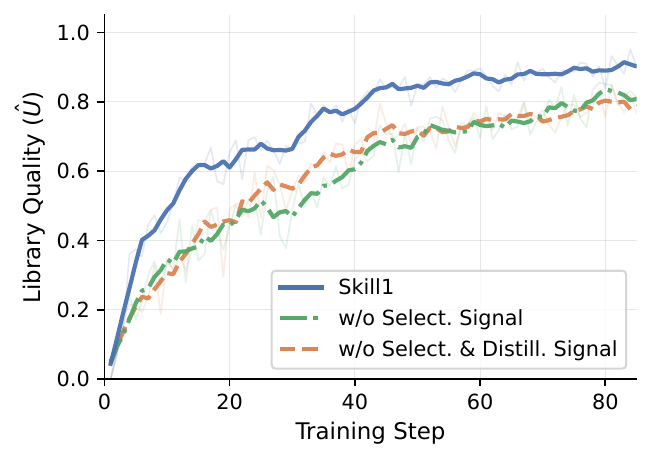}
        \caption{
        Top-skill utility ($\hat{U}$) during training. The variation signal drives the policy to distill increasingly effective skills.
        }
        \label{fig:library_quality}
    \end{minipage}
    \vspace{-0.5em}
\end{figure}

\textbf{The policy learns to generate increasingly precise selection queries.}
Figure~\ref{fig:retrieval_quality} measures task-skill similarity at three checkpoints.
Full Skill1 improves from 0.51 to 0.60 across training because the trend signal rewards queries that retrieve historically high-utility skills, gradually sharpening the policy's ability to describe what it needs.
Removing the selection signal slows this learning, and without learned selection entirely, similarity stays almost flat at the lowest level.

\textbf{The library ceiling rises as the policy learns to distill better skills.}
Figure~\ref{fig:library_quality} tracks $\hat{U}$, the utility of the top-ranked skill per task.
A rising $\hat{U}$ means increasingly effective skills are entering the library, not merely more skills.
Full Skill1 reaches 0.91 by step 85 while both ablations lag by approximately 0.10.
The variation signal creates this pressure: producing a skill similar to existing ones yields little reward, so the policy must discover genuinely better strategies to obtain positive gradient.

\subsubsection{Skill Library Diversity}
\label{sec:library_diversity}

\begin{figure}[t]
    \centering
    \includegraphics[width=\linewidth]{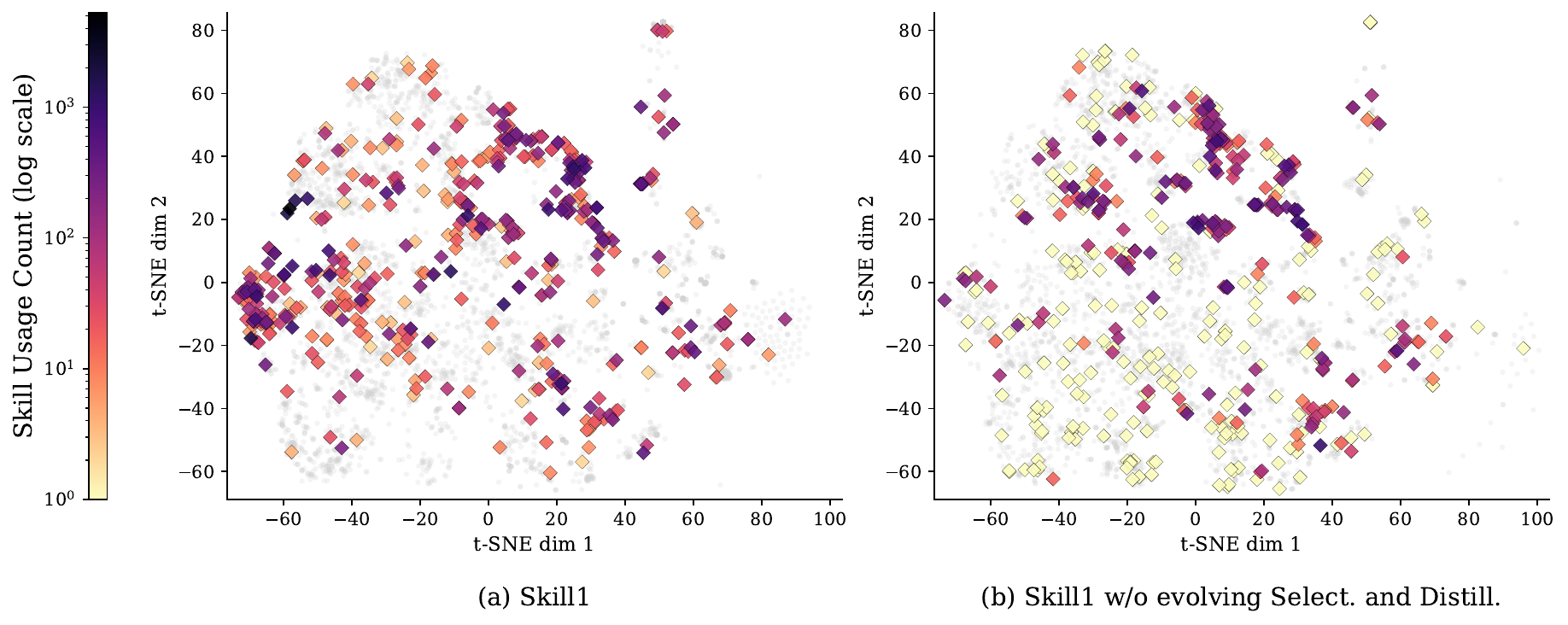}
    \caption{
    T-SNE visualization of the skill libraries after convergence, with and without RL-trained selection and distillation.
    The top-10 percent most frequently used skills are highlighted.
    Skill1 activates nearly twice as many high-frequency skills, and these skills span a broader strategy space.
    }
    \label{fig:skill_diversity}
\end{figure}

We examine whether the library is utilized as a diverse collective asset or collapses to a few dominant entries. Figure~\ref{fig:skill_diversity} visualizes the converged libraries with and without credit-assignment signals.

\textbf{Co-evolution activates a broader set of skills.}
Skill1 uses a broader set of skills.
As observed in Figure \ref{fig:skill_diversity}, the skill usage count distributes more uniformly in the left panel.
Without evolving signals (\ie Skill1 w/o Select. and Distill.), the skill usage count distribution sharpens, where only a small number of popular skills are intensively utilized.

\textbf{Frequently used skills cover diverse strategies.}
We also observe that the active skills in Skill1 span a much broader region of the strategy space.
On the contrary, the popular skills (red and purple ones) on the right subfigure huddle together with only limited coverage.
In the design of our method, producing a under-performing skill similar to existing ones yields negative reward, so the policy is pressured to cover underserved scenarios rather than duplicating successful ones.

\subsubsection{Computational Overhead}
\label{sec:compute}

We compare wall-clock time and library size for Skill1, SkillRL, and two ablations under identical hardware of 8 H800 80GB GPUs.

\begin{table}[t]
    \centering
    \small
    \caption{
    Computational cost on ALFWorld training. We report wall-clock time per step (seconds) and library size (number of skills) at three checkpoints.
    }
    \vspace{.2em}
    \begin{tabular}{l|ccc|ccc}
        \toprule
         &
        \multicolumn{3}{c|}{\textbf{Time / Step (s)}} &
        \multicolumn{3}{c}{\textbf{Library Size}} \\
        \cmidrule(lr){2-4} \cmidrule(lr){5-7}
        \textbf{Method} & Step 20 & Step 60 & Step 100 & Step 20 & Step 60 & Step 100 \\
        \midrule
        GRPO (no library) & 301.3 & 274.1 & 296.7 & --- & --- & --- \\
        SkillRL & 368.1 & 319.0 & 326.6 & 60  & 71 & 83 \\
        \midrule
        Skill1              & 386.6 & 444.3 & 493.8 & 915  & 3,899 & 5,000 \\
        \quad w/o Select. & 367.4 & 406.7 & 521.8 & 892  & 3,693 & 5,000 \\
        \quad w/o Distill. & 508.8 & 750.1 & 738.4 & 2,212  & 5,000 & 5,000 \\
        \bottomrule
    \end{tabular}
    \vspace{-.5em}
    \label{tab:compute}
\end{table}

\textbf{Skill1 adds moderate overhead over baseline methods.}
GRPO without a library runs at approximately 290s per step.
SkillRL maintains near-constant cost because its library grows minimally from 60 to 83 skills, but this static library limits final performance to 89.9\% compared to 97.5\% for Skill1.
Skill1 operates at 387 to 494s, roughly 1.3 to 1.7 times as slow as GRPO, with the increase stemming from the growing library context.
The selection step itself adds negligible overhead as query generation and re-ranking operate on short sequences compared to multi-turn interactions against the environment.

\textbf{Distillation controls both library quality and computational cost.}
Without distillation, raw trajectories enter the library directly, growing it at 2.4 times the rate of Skill1.
The larger library lengthens the selection context, making the variant without distillation 69\% slower by step 60 and saturating the 5,000-skill cap far earlier.
Distillation compresses experience into concise skills, improving library quality while keeping computational cost in check.

%% file: chapters/5_conclusion.tex


\section{Conclusion and Limitations}
\label{sec:conclusion}

\paragraph{Conclusion.}
We present Skill1, a framework that trains a single policy to co-evolve skill selection, utilization, and distillation toward a shared task-outcome objective.
By decomposing this signal into its low-frequency trend and high-frequency variation, Skill1 derives per-capability credit assignment without auxiliary rewards.
Experiments on ALFWorld and WebShop show consistent gains over prior skill-based and RL baselines, and ablations confirm that the three capabilities evolve in a coupled manner.
A natural next step is to extend this lifecycle to hierarchical or multi-agent settings, where skill sharing and conflict resolution introduce new challenges for unified credit assignment.

\paragraph{Limitations.}
\label{sec:limitations}
While Skill1 achieves strong performance, several limitations remain.
\begin{itemize}[leftmargin=*]
    \item \textbf{Environment coverage.}
    Our evaluation is limited to two representative text-based agent environments.
    Whether the co-evolution framework generalizes to more environments (\eg deep search environments) or those with visual observations remains unexplored.
    \item \textbf{Scalability of the skill library.}
    The library capacity in this work is capped at 5,000 entries.
    As the diversity of tasks grows, the fixed-size library may become a bottleneck, and more sophisticated eviction or hierarchical organization strategies may be required.
\end{itemize}

%% file: chapters/99_appendix.tex
\newpage
\appendix

\input{chapters/9_related_work}

\section{Algorithm Details}
\label{app:grpo}

We use Group Relative Policy Optimization (GRPO)~\citep{deepseekmath} as the optimization method, which eliminates the need for a separate value network by computing advantages relative to a group of rollouts sampled from the same task.
For each task $d$, a group of $G$ rollouts $\{\tau_i\}_{i=1}^{G}$ is sampled from $\pi_{\theta_\text{old}}$.
The group-relative advantage for rollout $i$ is:
\begin{equation}
\label{eq:grpo_advantage}
\hat{A}_i = \frac{r(\tau_i) - \operatorname{mean}(\{r(\tau_1), \ldots, r(\tau_G)\})}{\operatorname{std}(\{r(\tau_1), \ldots, r(\tau_G)\})}.
\end{equation}
Let $\rho_t^{(i)}(\theta) = \pi_\theta(a_t^{(i)} \mid s_t^{(i)}) / \pi_{\theta_\text{old}}(a_t^{(i)} \mid s_t^{(i)})$ denote the per-token importance ratio.
The GRPO objective maximizes the clipped surrogate:
\begin{equation}
\label{eq:grpo}
\mathcal{J}_\text{GRPO}(\theta) = \frac{1}{G} \sum_{i=1}^{G} \frac{1}{|\tau_i|} \sum_{t=1}^{|\tau_i|} \min\!\bigl(\rho_t^{(i)} \hat{A}_i,\; \operatorname{clip}(\rho_t^{(i)}, 1{-}\epsilon, 1{+}\epsilon)\, \hat{A}_i\bigr) - \beta\, D_\text{KL}\!\bigl[\pi_\theta \| \pi_\text{ref}\bigr],
\end{equation}
where $\epsilon$ is the clipping ratio, $\beta$ controls KL regularization toward a reference policy $\pi_\text{ref}$, and $|\tau_i|$ is the number of tokens in rollout $i$.

\section{Implementation Details}
\label{app:hyperparams}

\paragraph{Training infrastructure.}
Skill1 is trained on 8 NVIDIA H800-80GB GPUs using the VeRL framework~\citep{verl} with Fully Sharded Data Parallelism (FSDP) under BFloat16 precision.
Rollout generation uses vLLM with tensor parallelism of 4.
Training converges in approximately 100 to 150 steps (roughly 30 hours on ALFWorld).
The auxiliary objective weights are $\lambda_1 = \lambda_2 = 0.3$ throughout all experiments unless otherwise specified.

\paragraph{Baseline reproduction.}
We reproduce RetroAgent using its official implementation.\footnote{\url{https://github.com/zhangxy-2019/RetroAgent}}
For SkillRL, EvolveR, Mem0, and SimpleMem, we use numbers reported in their respective papers~\citep{skillrl, evolver, mem0, simplemem} under the same base model (Qwen2.5-7B-Instruct).
GiGPO results are taken from~\citet{gigpo}.
All RL baselines use identical training budgets (150 epochs) and the same train/test splits to ensure fair comparison.

\paragraph{Hyperparameters.}
Table~\ref{tab:hyperparams_shared} lists the shared training hyperparameters across both environments.
Table~\ref{tab:hyperparams_env} lists the per-environment differences.
Table~\ref{tab:hyperparams_skill} lists the skill library configuration.

\begin{table}[ht]
    \centering
    \small
    \caption{Shared training hyperparameters.}
    \begin{tabular}{lc}
        \toprule
        \textbf{Hyperparameter} & \textbf{Value} \\
        \midrule
        \multicolumn{2}{l}{\textit{Optimization}} \\
        Algorithm & GRPO \\
        Learning rate & $1 \times 10^{-6}$ \\
        KL loss coefficient & 0.01 \\
        KL loss type & low-variance KL \\
        PPO mini-batch size & 256 \\
        PPO micro-batch size per GPU & 16 \\
        Gradient checkpointing & True \\
        Re-ranking loss weight $\lambda_1$ & 0.3 \\
        Distillation loss weight $\lambda_2$ & 0.3 \\
        \midrule
        \multicolumn{2}{l}{\textit{Rollout}} \\
        Group size $G$ & 16 \\
        Max prompt length & 16,384 tokens \\
        Max response length & 2,048 tokens \\
        vLLM tensor parallelism & 4 \\
        GPU memory utilization & 0.7 \\
        Validation temperature & 0.4 \\
        \bottomrule
    \end{tabular}
    \label{tab:hyperparams_shared}
\end{table}

\begin{table}[ht]
    \centering
    \small
    \caption{Per-environment hyperparameters.}
    \begin{tabular}{lcc}
        \toprule
        \textbf{Hyperparameter} & \textbf{ALFWorld} & \textbf{WebShop} \\
        \midrule
        Training batch size & 16 & 32 \\
        Validation batch size & 64 & 128 \\
        Max environment steps & 50 & 15 \\
        \bottomrule
    \end{tabular}
    \label{tab:hyperparams_env}
\end{table}

\begin{table}[ht]
    \centering
    \small
    \caption{Skill library configuration.}
    \begin{tabular}{lc}
        \toprule
        \textbf{Parameter} & \textbf{Value} \\
        \midrule
        \multicolumn{2}{l}{\textit{Selection}} \\
        Encoder & all-MiniLM-L6-v2 (384-dim) \\
        Top-$K$ candidates & 5 \\
        Training selection strategy & UCB \\
        Evaluation selection strategy & Greedy \\
        UCB exploration scale & 1.0 \\
        Similarity weight $w_\text{sim}$ & 0.6 \\
        \midrule
        \multicolumn{2}{l}{\textit{Library Management}} \\
        Maximum library size & 5,000 \\
        Utility EMA rate $\alpha$ & 0.05 \\
        \bottomrule
    \end{tabular}
    \label{tab:hyperparams_skill}
\end{table}

\section{Statistical Analysis}
\label{app:statistical_analysis}


We run all methods with 3 independent random seeds and report mean $\pm$ standard deviation (1-$\sigma$).
The primary source of variability is the random seed, which affects parameter initialization, rollout sampling order, and skill library evolution trajectory.
We use SciPy's \texttt{ttest\_ind} with \texttt{equal\_var=False} (Welch's t-test) to assess statistical significance.

\subsection{Full Performance Breakdown}

\begin{figure}[h]
    \centering
    \includegraphics[width=0.85\linewidth]{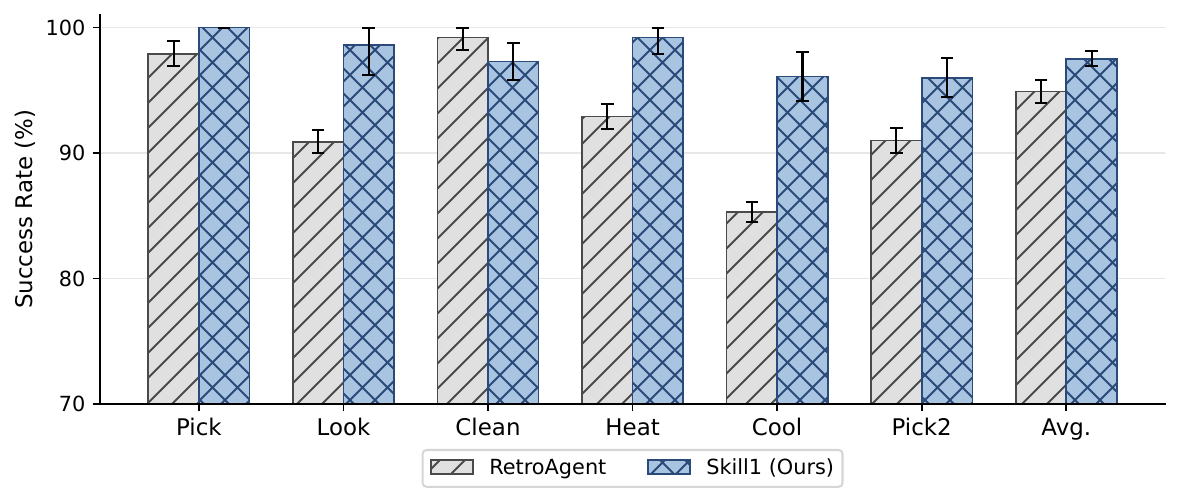}
    \caption{Per-task success rates (mean $\pm$ std over 3 seeds). Skill1 consistently outperforms best baseline RetroAgent on five of six task types and the average score.}
    \label{fig:error_bar}
\end{figure}

We select RetroAgent as the strongest baseline and run it with 3 independent seeds under identical conditions to obtain variance estimates.
Figure~\ref{fig:error_bar} reports per-task-type success rates (mean $\pm$ std) on ALFWorld.

\subsection{Analysis}

\textbf{Skill1 achieves statistically significant improvement over RetroAgent.}
On the aggregate metric (ALF All), Skill1 achieves 97.5$\pm$0.6 versus RetroAgent's 94.9$\pm$0.9.
A Welch's t-test on the 3-seed averages yields $t = 4.06$, $\mathrm{df} = 3.40$, $p = 0.021$ ($< 0.05$).
The result confirms that the gain is not attributable to seed variance.
Per-task significance is strongest on the tasks where RetroAgent struggles most: Heat ($p = 0.004$), Cool ($p = 0.005$), and Look ($p = 0.020$).
The sole exception is Clean, where Skill1 trails RetroAgent by 1.9 points. This difference is not statistically significant ($p = 0.147 > 0.05$) and falls within normal seed variance.

\textbf{Skill1 exhibits lower aggregate variance than RetroAgent.}
Skill1's overall standard deviation (0.6) is smaller than RetroAgent's (0.9), indicating more stable convergence across seeds.
The unified evolution framework, where selection, utilization, and distillation reinforce each other, reduces sensitivity to initialization.

\section{Broader Impacts}
\label{sec:broader_impacts}


This work develops a framework for LLM agents to autonomously acquire and reuse behavioral skills through reinforcement learning.
On the positive side, the approach can reduce the manual engineering effort required to build capable agents and enable more sample-efficient learning in interactive environments.

On the negative side, agents that autonomously accumulate skills may exhibit emergent behaviors that are difficult to predict or audit. In high-stakes deployment scenarios, an unconstrained skill library could encode harmful action sequences, and brings new injection risks. We recommend deploying such systems with human-in-the-loop oversight and constraining the action space in safe domains.

\section{Case Studies}
\label{app:case_studies}

We present two representative case studies from the ALFWorld evaluation, comparing Skill1 against RetroAgent on the same test task.
Each case demonstrates a different transfer mechanism (failure avoidance and error correction) and highlights why unified evolution of selection, utilization, and distillation produces qualitatively different behavior from baselines that lack joint optimization.

\definecolor{casebg}{RGB}{246, 248, 252}
\definecolor{caseborder}{RGB}{140, 155, 180}
\definecolor{skillbg}{RGB}{240, 244, 250}
\definecolor{skillborder}{RGB}{120, 140, 170}


\begin{tcolorbox}[colback=casebg, colframe=caseborder, title={\textbf{Case Study 1: Failure Avoidance via Appliance Selection}}, fonttitle=\bfseries\small, boxrule=0.6pt, arc=2mm, left=4pt, right=4pt, top=3pt, bottom=3pt]
\small

\textbf{Test task:} \texttt{heat some plate and put it in cabinet.}

\textbf{Challenge:} The environment contains both stoveburners and a microwave. Stoveburners are non-functional for plates, but this constraint is invisible from the task description.

\tcblower

\vspace{4pt}

\begin{tcolorbox}[colback=skillbg, colframe=skillborder, boxrule=0.4pt, arc=1.5mm, left=3pt, right=3pt, top=2pt, bottom=2pt, title={\small Skill1: Retrieved Skill (Utility: 0.951, Used 593 times)}, fonttitle=\small]
\footnotesize
\textbf{Origin task:} \texttt{heat some plate and put it in fridge.}\\
\textbf{Description:} Useful when the agent needs to heat an object using a microwave, especially when stove burners are not effective. \\
\textbf{Strategy:} ``Successfully used microwave 1 to heat the plate. The attempt failed to heat the plate using the stoveburners, which might not have been operational. The microwave 1 was found on countertop 2.''
\end{tcolorbox}

\vspace{4pt}
\textbf{Skill1 reasoning (step 3):} ``I have already picked up the plate from countertop 1. \textbf{The microwave 1 is the most suitable option for heating the plate, as mentioned in past reflections.} I will go to the microwave 1 to heat the plate.''

\vspace{4pt}
\textbf{Skill1 execution trace (6 actions, optimal):}
\begin{enumerate}[nosep, leftmargin=1.5em]
\item \texttt{go to countertop 1}
\item \texttt{take plate 1 from countertop 1}
\item \colorbox{lightgreen}{\texttt{go to microwave 1}} 
\item \colorbox{lightgreen}{\texttt{heat plate 1 with microwave 1}}
\item \texttt{go to cabinet 1}
\item \texttt{put plate 1 in cabinet 1} ~~\textcolor{green!60!black}{\ding{51} SUCCESS}
\end{enumerate}

\end{tcolorbox}

\vspace{6pt}


\begin{tcolorbox}[colback=casebg, colframe=caseborder, title={\textbf{Case Study 2: Error Correction via Object Targeting}}, fonttitle=\bfseries\small, boxrule=0.6pt, arc=2mm, left=4pt, right=4pt, top=3pt, bottom=3pt]
\small

\textbf{Test task:} \texttt{examine the alarmclock with the desklamp.}

\textbf{Challenge:} The room contains multiple alarmclock instances. Targeting the wrong instance leads to task failure.

\tcblower

\vspace{4pt}

\begin{tcolorbox}[colback=skillbg, colframe=skillborder, boxrule=0.4pt, arc=1.5mm, left=3pt, right=3pt, top=2pt, bottom=2pt, title={\small Skill1: Retrieved Skill (Utility: 0.997, Used 112 times)}, fonttitle=\small]
\footnotesize
\textbf{Origin task:} \texttt{look at alarmclock under the desklamp.}\\
\textbf{Description:} Useful when the agent needs to target a specific object under a desklamp before using the desklamp to view the object.\\
\textbf{Strategy:} ``The desklamp 1 was used on alarmclock 2 instead of the alarmclock under the desklamp. The agent should locate the alarmclock under the desklamp first before using the desklamp.''
\end{tcolorbox}

\vspace{4pt}
\textbf{Skill1 reasoning (step 1):} ``I need to find an alarmclock and use the desklamp on it. I'll start by examining the dresser 1 to find both the alarmclock and desklamp.''

\vspace{4pt}
\textbf{Skill1 execution trace (3 actions, minimal):}
\begin{enumerate}[nosep, leftmargin=1.5em]
\item \texttt{go to dresser 1}
\item \colorbox{lightgreen}{\texttt{take alarmclock 1 from dresser 1}}
\item \texttt{use desklamp 1} ~~\textcolor{green!60!black}{\ding{51} SUCCESS}
\end{enumerate}

\end{tcolorbox}

\vspace{6pt}

\noindent\textbf{Discussion.}
Both cases illustrate how the co-evolved skill library captures knowledge that goes beyond surface-level pattern matching.
Case~1 demonstrates failure avoidance: RetroAgent lacks a high-utility skill encoding the stoveburner constraint because its selection mechanism is not optimized to route heat-tasks to the relevant skill. Skill1 retrieves the correct skill and explicitly cites it in its reasoning chain.
Case~2 demonstrates error correction: RetroAgent picks the wrong alarmclock instance because its library does not preserve the targeting lesson from prior failures with sufficient utility. Skill1's variation-driven distillation retains such lessons and the trend-driven selection surfaces them at test time.
In both cases, Skill1 achieves near-optimal trajectories while the baseline exhausts steps on avoidable mistakes.

\section{Prompt Templates}
\label{app:prompts}

\definecolor{promptblue}{RGB}{230, 240, 255}
\definecolor{promptborder}{RGB}{100, 140, 200}
\definecolor{promptgreen}{RGB}{240, 247, 248}
\definecolor{promptgreenborder}{RGB}{120, 160, 165}
\definecolor{promptorange}{RGB}{247, 244, 240}
\definecolor{promptorangeborder}{RGB}{165, 150, 130}

We list the prompt templates used in each stage of Algorithm~\ref{alg:skill1}:

\begin{itemize}[nosep, leftmargin=1.5em]
\item \textbf{Selection (Query generation)} (line 3): $\pi_\theta$ generates query $q$ to retrieve candidates from $\mathcal{B}$.
\item \textbf{Selection (Re-ranking)} (line 6): $\pi_\theta$ ranks $\mathcal{B}_K$ and selects the top skill $z$.
\item \textbf{Utilization} (line 8): $\pi_\theta$ interacts with the environment conditioned on $z.\text{strat}$.
\item \textbf{Distillation} (line 9): $\pi_\theta$ reflects on $\tau$ and produces $s_\text{new}$.
\end{itemize}

\subsection{ALFWorld}
\label{app:prompt_alfworld}

\begin{tcolorbox}[colback=promptgreen, colframe=promptgreenborder, title={Query Generation}, fonttitle=\bfseries\small, boxrule=0.6pt, arc=2mm]
\small
Task: \{TASK\}\\
Observation: \{INITIAL\_OBSERVATION\}

Write a one-sentence search query to find relevant past experiences for this task. Do NOT output an action.

Example: \verb|<query>|tips for heating an object with microwave then placing it\verb|</query>|

\verb|<query>|
\end{tcolorbox}

\begin{tcolorbox}[colback=promptgreen, colframe=promptgreenborder, title={Re-ranking}, fonttitle=\bfseries\small, boxrule=0.6pt, arc=2mm]
\small
You are about to attempt a task in the ALFRED Embodied Environment.

Task: \{TASK\}\\
Initial Observation: \{INITIAL\_OBSERVATION\}

Below are \{$K$\} past experiences retrieved from memory. Each is labeled with an ID.

\{CANDIDATE\_EXPERIENCES\}

Rank these experiences from MOST useful to LEAST useful for the current task. Consider which experience addresses the specific challenges you expect to face.

Output ONLY the ranked IDs as a comma-separated list within \verb|<rank>| \verb|</rank>| tags.
\end{tcolorbox}

\begin{tcolorbox}[colback=promptgreen, colframe=promptgreenborder, title={Utilization}, fonttitle=\bfseries\small, boxrule=0.6pt, arc=2mm]
\small
You are an expert agent operating in the ALFRED Embodied Environment. Your task is to: \{TASK\}

\textcolor{gray}{\textit{[Injected if a skill is selected:]}}\\
\textcolor{gray}{Past reflections on similar tasks: \{SKILL.strat\}}\\
\textcolor{gray}{Warning: These lessons may be outdated. Use them only if they align with your current observation.}

Prior to this step, you have already taken \{N\} step(s). Below are the most recent \{W\} observations and the corresponding actions you took: \{ACTION\_HISTORY\}

You are now at step \{CURRENT\_STEP\} and your current observation is: \{OBSERVATION\}\\
Your admissible actions of the current situation are: [\{ADMISSIBLE\_ACTIONS\}].

You should first reason step-by-step within \verb|<think>| \verb|</think>| tags. Then choose an admissible action within \verb|<action>| \verb|</action>| tags.
\end{tcolorbox}

\begin{tcolorbox}[colback=promptgreen, colframe=promptgreenborder, title={Distillation}, fonttitle=\bfseries\small, boxrule=0.6pt, arc=2mm]
\small
You are an expert evaluating an ALFRED Embodied Environment task attempt.\\
Your task is to: \{TASK\}\\
The task was \{successfully/unsuccessfully\} completed.

Trajectory of the attempt: \{TRAJECTORY\}

\verb|<think>| Analyze: What subtasks were attempted (pick up, navigate, use appliance, place)? Which succeeded or failed? What specific actions led to this outcome? What is the most valuable lesson? \verb|</think>|

Output your evaluation as JSON:\\
\verb|{"task_success": ..., "action_lesson": "...", "navigation_lesson": "...",|\\
\verb| "description_head": "[WHEN this lesson is useful -- general task type, not specific task]"}|
\end{tcolorbox}

\subsection{WebShop}
\label{app:prompt_webshop}

\begin{tcolorbox}[colback=promptorange, colframe=promptorangeborder, title={Query Generation}, fonttitle=\bfseries\small, boxrule=0.6pt, arc=2mm]
\small
Task: \{TASK\}\\
Observation: \{INITIAL\_OBSERVATION\}

Write a one-sentence search query to find relevant past experiences for this task. Do NOT output an action.

Example: \verb|<query>|tips for finding products with specific color and size under budget\verb|</query>|

\verb|<query>|
\end{tcolorbox}

\begin{tcolorbox}[colback=promptorange, colframe=promptorangeborder, title={Re-ranking}, fonttitle=\bfseries\small, boxrule=0.6pt, arc=2mm]
\small
You are about to attempt a shopping task in the WebShop environment.

Task: \{TASK\}\\
Initial Observation: \{INITIAL\_OBSERVATION\}

Below are \{$K$\} past experiences retrieved from memory. Each is labeled with an ID.

\{CANDIDATE\_EXPERIENCES\}

Rank these experiences from MOST useful to LEAST useful for the current task. Consider which experience addresses the specific challenges you expect to face.

Output ONLY the ranked IDs as a comma-separated list within \verb|<rank>| \verb|</rank>| tags.
\end{tcolorbox}

\begin{tcolorbox}[colback=promptorange, colframe=promptorangeborder, title={Utilization}, fonttitle=\bfseries\small, boxrule=0.6pt, arc=2mm]
\small
You are an expert autonomous agent operating in the WebShop e-commerce environment.

\textcolor{gray}{\textit{[Injected if a skill is selected:]}}\\
\textcolor{gray}{Past reflections on similar tasks: \{SKILL.strat\}}\\
\textcolor{gray}{Warning: These lessons may be outdated. Use them only if they align with your current situation.}

Your task is to: \{TASK\}.\\
Prior to this step, you have already taken \{N\} step(s). Below are the most recent \{W\} observations and the corresponding actions you took: \{ACTION\_HISTORY\}

You are now at step \{CURRENT\_STEP\} and your current observation is: \{OBSERVATION\}.\\
Your admissible actions: [\{AVAILABLE\_ACTIONS\}].

You should first reason step-by-step within \verb|<think>| \verb|</think>| tags, then choose an admissible action within \verb|<action>| \verb|</action>| tags.
\end{tcolorbox}

\begin{tcolorbox}[colback=promptorange, colframe=promptorangeborder, title={Distillation}, fonttitle=\bfseries\small, boxrule=0.6pt, arc=2mm]
\small
You are an expert evaluating a WebShop shopping attempt.\\
Your task is to: \{TASK\}\\
The task was \{successfully/unsuccessfully\} completed.

Trajectory of the attempt: \{TRAJECTORY\}

\verb|<think>| Analyze: What subtasks were attempted (search, filter, select, purchase)? Which succeeded or failed? What specific actions led to this outcome? What are the most valuable lessons? \verb|</think>|

Output your evaluation as JSON:\\
\verb|{"task_success": ..., "action_lesson": "...", "navigation_lesson": "...",|\\
\verb| "description_head": "[WHEN this lesson is useful -- general task type, not specific task]"}|
\end{tcolorbox}

%% file: chapters/9_related_work.tex
\section{Related Work}
\label{sec:related_work}


\paragraph{Reinforcement Learning for LLM Agents.}
Core algorithmic advances include GRPO~\citep{deepseekmath}, anchor-state grouping~\citep{gigpo}, and dynamic sampling with asymmetric clipping~\citep{dapo}.
Multi-turn RL methods address long-horizon challenges through hierarchical value functions~\citep{archer}, leave-one-out advantage estimation~\citep{loop}, MCTS-guided search~\citep{agent-q}, exploration-based trajectory optimization~\citep{eto}, multi-turn self-evolution~\citep{ragen,agentrl}, and cross-episode meta-RL~\citep{lamer}.
Recent work further refines credit assignment via stepwise progress attribution~\citep{spa-rl,turn-level-reward} or intrinsic exploration signals~\citep{intrinsic-motivation-llm,entropy-modulated}.
Prompt-based methods such as ReAct~\citep{react} and Reflexion~\citep{reflexion} enable reasoning without parameter updates but are upper-bounded by the frozen policy~\citep{continual-rl}.
Skill1 extends GRPO by decomposing a single task-outcome signal into stage-specific gradients for selection, utilization, and distillation within a unified RL objective.

\paragraph{Experience Reusing.}
Structuring past experience for reuse improves RL sample efficiency~\citep{exgrpo,oel,comp-rl}, and explicit memory systems that store interaction histories~\citep{evo-memory,simplemem,exploratory-memory} or distilled lessons~\citep{memp,memento,cogito} support continuous adaptation.
RetroAgent~\citep{retroagent} combines intrinsic progress rewards with language-based lesson extraction and a utility-aware selection strategy~\citep{ucb}.
Critique-GRPO~\citep{critique-grpo} integrates natural-language critiques with numerical rewards, and RL-based self-distillation~\citep{rl-self-distillation} refines failed trajectories into policy updates.
Retrospective self-correction through natural-language critiques~\citep{self-refine,retroformer} further enables agents to learn from failures~\citep{self-improving-position}.
Skill1 builds on these insights but derives all learning signals from a single task-outcome signal, eliminating the need for separate intrinsic reward design.

\paragraph{Skill Libraries for LLM Agents.}
A growing body of work equips LLM agents with persistent skill libraries~\citep{sok-agentic-skills,agent-skills-survey,skill-ecosystem,xskill,anthropic-skills}.
For selection, approaches include frozen embedding selectors~\citep{skillrl,comp-rl}, heuristic scoring~\citep{retroagent}, learned routing~\citep{memskill,skillorchestra}, and policy log-probability ranking~\citep{arise,evolver}.
For utilization, RL-based methods condition the policy on selected skills~\citep{skillrl,comp-rl,retroagent,arise,sage}, sometimes with hierarchical rewards to incentivize skill use~\citep{arise,comp-rl}.
For distillation, methods range from prompt-based extraction~\citep{expel} and training-free skill versioning~\citep{autoskill} to teacher-driven generation~\citep{skillrl}, co-evolving extractors~\citep{comp-rl}, and self-reflection~\citep{retroagent,sage,evolver,mia}.
Existing methods have not yet achieved RL-optimized status on all three stages simultaneously, and those that optimize multiple stages use heterogeneous learning signals without a unified objective.
Skill0~\citep{skill0} internalizes skills into model parameters with zero external skills; Skill1 co-evolves all three stages through one policy model and a unified task outcome signal.